\documentclass[letterpaper, 10 pt, conference]{ieeeconf}  %

\IEEEoverridecommandlockouts                              %

\usepackage{graphics} %
\usepackage{hyperref}
\usepackage[separate-uncertainty = true,multi-part-units=single]{siunitx}
\usepackage[sortcites,backend=biber,hyperref=true]{biblatex}
\bibliography{bibliography}
\usepackage{tikz}
\usetikzlibrary{3d, angles, calc, quotes}

\newcommand*{\Nm}[1]{\SI{#1}{\newton\metre}}

\def\robotName{Athena}
\def\prevRobotName{Asterix}
\author{Stefan Fabian, Aljoscha Schmidt, Jonas Süß, Dishant, Aum Oza and Oskar von Stryk%
 \thanks{Stefan Fabian, Aljoscha Schmidt and Oskar von Stryk are with the Technical University of Darmstadt,
     Computer Science Department, Simulation, Systems Optimization and Robotics Group, Germany.\newline
     {\tt\small \{fabian,schmidt,stryk\}@sim.tu-darmstadt.de}}%
 \thanks{%
     Research presented in this paper has been supported in parts by the LOEWE initiative (Hesse, Germany) within the emergenCITY center [LOEWE/1/12/519/03/05.001(0016)/72] and by the German Federal Ministry of Education and Research (BMBF) within the DRZ project (grant no. 13N16475) and the RIG project (grant no. 16ME1001).%
     \newline 979-8-3315-4526-0/25/\$31.00 \textcopyright 2025 IEEE%
     \newline DOI: 10.1109/SSRR68451.2025.11391264}%
}

\title{\vspace{-2cm}%
\small Preprint of the paper which appeared in:\\ \large IEEE International Symposium on Safety, Security, and Rescue Robotics (SSRR) 2025%
\vspace{0.5cm}\newline\LARGE\bf%
\robotName: An Autonomous Open-Hardware Tracked Rescue Robot Platform
}

\begin{document}

\maketitle
\thispagestyle{empty}
\pagestyle{empty}

\begin{abstract}%
In disaster response and situation assessment, robots have great potential in reducing the risks to the safety and health of first responders.
As the situations encountered and the required capabilities of the robots deployed in such missions differ wildly and are often not known in advance, heterogeneous fleets of robots are needed to cover a wide range of mission requirements.
While UAVs can quickly survey the mission environment, their ability to carry heavy payloads such as sensors and manipulators is limited.
UGVs can carry required payloads to assess and manipulate the mission environment, but need to be able to deal with difficult and unstructured terrain such as rubble and stairs.
The ability of tracked platforms with articulated arms (flippers) to reconfigure their geometry makes them particularly effective for navigating challenging terrain.
In this paper, we present \robotName, an open-hardware rescue ground robot research platform with four individually reconfigurable flippers and a reliable low-cost remote emergency stop (E-Stop) solution.
A novel mounting solution using an industrial PU belt and tooth inserts allows the replacement and testing of different track profiles.
The manipulator with a maximum reach of 1.54m can be used to operate doors, valves, and other objects of interest.
Full CAD \& PCB files, as well as all low-level software, are released as open-source contributions.
\end{abstract}

\section{INTRODUCTION}

The response to disasters poses a high risk to the health and safety of first responders.
In situations where a response is required but very dangerous and not time-critical, e.g., in the case of explosive ordnance disposal or inspection of buildings at risk of collapse, robots can and are increasingly used to protect human lives and reduce exposure to hazardous environments.

The first documented use of robots in disaster response dates back to 1986 in Chernobyl, where robots were used for radiation reconnaissance and cleanup~\cite{bishop1987robots}.
Since then, robots have been used successfully in numerous disaster response scenarios, such as the collapse of the World Trade Center in 2001, the Fukushima nuclear disaster in 2011, and many more~\cite{matsuno2013utilization}.
Advances in robotics, such as the development of more capable and affordable sensors, better computing hardware, and more powerful actuation hardware, have enabled the development of more capable robots that can intelligently assist and enhance human disaster response teams.

As disaster response missions are, by definition, unpredictable, occurring in unknown and often dangerous and unstructured environments, rescue robots need to be highly flexible and adaptable in their capabilities.
While some missions may only require a simple inspection of the environment, others may require the manipulation of objects, such as opening doors or valves, or even the removal of debris.
Additionally, the robot may need to be able to traverse unstructured terrain and obstacles, such as rubble or stairs.
This unpredictability and the need for flexibility in responding to disasters pose a significant challenge for designing rescue robots.
As different robots have different strengths and weaknesses, a heterogeneous team of robots is often required to cover the full range of possible scenarios.
While UAVs are very effective for reconnaissance and inspection, they are limited in their ability to carry payloads and manipulate objects.
Legged robots such as \cite{hutter2017anymal} are highly mobile and can traverse a wide range of terrains, but they are also limited in their payload capacity and may encounter difficulties with unstructured and unstable terrain.
Tracked robots, on the other hand, are highly stable and can carry larger payloads, but are typically less agile than legged robots.

\begin{figure}
  \centering
  \includegraphics[width=\linewidth]{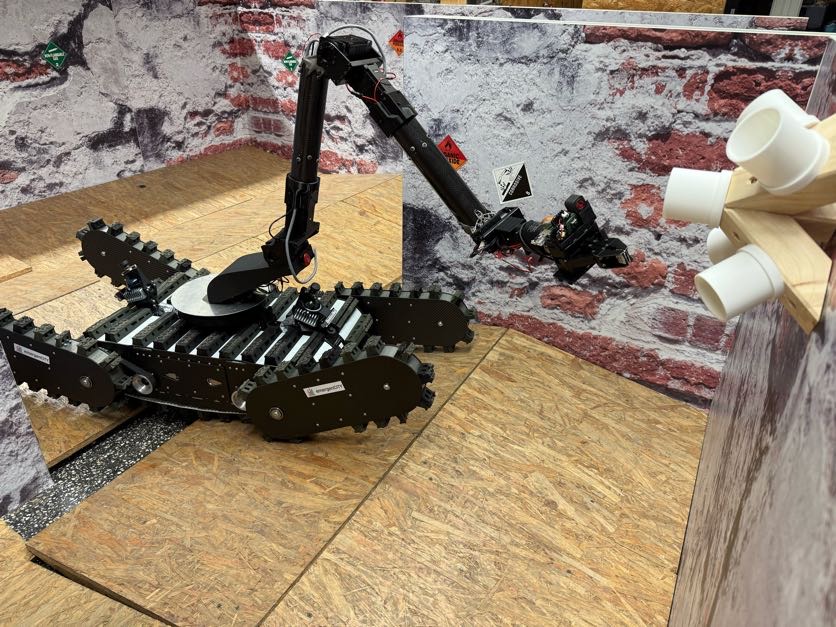}
  \caption{\robotName, a highly mobile tracked rescue robot with four individually movable flippers and a large manipulator arm.}
\end{figure}

While many different tracked designs exist~\cite{trackedlocomotionsystems2022}, we concentrate on tracked robots with actively reconfigurable track arms (flippers) that can adapt to unstructured terrain and obstacles.
Among these, several commercial platforms are available, including the Telerob Telemax, Jaguar V6, and Tready by HEBI Robotics.
However, these platforms are typically high-cost and closed designs that provide limited interfaces and hinder modifications, often necessary in research.
While there is extensive research on autonomy and assistance functions using articulated tracked robots, information on the hardware or fully open hardware designs is limited~\cite{edlinger2019marc,nagatani2011redesign,habibian2021design,asterix2021}.

In this paper, we present \robotName, a highly mobile modular tracked rescue robot with four individually movable flippers and an innovative belt design with exchangeable mounted profiles.
A novel combination of LiDAR and cameras provides high coverage of the robot's environment, and a large manipulator arm allows for the manipulation of mission-critical objects, such as collecting probes, manipulating valves, and buttons.
In the following section, we will introduce the locomotion system, followed by a detailed description of the manipulator design.
After introducing the mechanical parts, we present the electronics, including sensors, power distribution, compute, and our solution for a remote E-Stop
to improve hardware safety during testing.
Subsequently, the results of our evaluation at the RoboCup Rescue League as well as the evaluation of the robot's limits in step height and manipulator payload are presented, followed by an evaluation of the trigger speed of our remote E-Stop.
The main contributions of this paper are the full design of a lightweight (\SI{50}{\kilo\gram}), compact (\SI{0.7}{\metre}~x~\SI{0.5}{\metre}) and affordable  -- less than 60.000€ in materials -- tracked rescue robot well suited for manipulation and obstacle traversal research, including CAD / PCB files, and low-level software, and a low-cost (approx. 83€ in materials), reliable remote E-Stop sender and receiver for safe experiments\footnote{\url{https://github.com/tu-darmstadt-ros-pkg/athena}.}.

\section{RELATED WORK}
In the class of UGVs with manipulators, numerous designs for rescue robots exist, each with its strengths and weaknesses.
While heavy and highly robust robots, such as the Earthshaker, are capable of removing debris of up to \SI{50}{\kilogram} and driving through deep water puddles in outdoor terrains \cite{earthshaker2023}, their size and weight significantly impact their agility and ability to navigate tight spaces and overcome obstacles.

\begin{table*}[t]
\centering
\caption{Comparison of articulated tracked rescue robots with available published design papers.}
\renewcommand{\arraystretch}{1.2}
\begin{tabular}{l p{3.2cm} p{3cm} p{3cm} p{3.2cm}}
\hline
\textbf{Feature} & \textbf{Quince} \cite{rohmer2010quince} (revised \cite{nagatani2011redesign}) & \textbf{Karo} \cite{habibian2021design} & \textbf{Asterix} \cite{asterix2021} & \textbf{Athena} (ours)\\
\hline
\multicolumn{5}{l}{\textbf{Locomotion}} \\
Flipper DoF & 4 (independent) & 2 (coupled pairs) & 2 (coupled pairs) & 4 (independent) \\
Maximum Speed & \SI[per-mode=symbol]{1.1}{\meter\per\second} & \SI[per-mode=symbol]{0.8}{\meter\per\second} & \SI[per-mode=symbol]{1.1}{\meter\per\second} & \SI[per-mode=symbol]{1.5}{\meter\per\second} \\
Maximum Step Height & \SI{34}{\centi\meter} & N/A & \SI{30}{\centi\meter} & \SI{41}{\centi\meter}\\

\multicolumn{5}{l}{\textbf{Manipulation}} \\
Manipulator Arm DoF & 0 (2\textsuperscript{*}) & 6  & 6 & 7 \\
Maximum Reach & N/A & \SI{1.3}{\meter} & \SI{0.85}{\meter} & \SI{1.54}{\meter} \\
Payload (at max. reach) & N/A & \SI{5.0}{\kilogram} & \SI{2.5}{\kilogram} & \SI{2.9}{\kilogram} \\[3pt]

\multicolumn{5}{l}{\textbf{General}} \\
Weight & \SI{27}{\kilogram} (\SI{50}{\kilogram}\textsuperscript{*}) & \SI{85}{\kilogram} & \SI{57.3}{\kilogram} & \SI{50}{\kilogram} \\
Dimensions (L $\times$ W) & \SI{0.71}{\meter}$\times$\SI{0.42}{\meter} & \SI{0.8}{\meter}$\times$\SI{0.6}{\meter} & \SI{0.71}{\meter}$\times$\SI{0.52}{\meter} & \SI{0.7}{\meter}$\times$\SI{0.5}{\meter} \\[3pt]

\multicolumn{5}{l}{\textbf{Sensors}} \\
 & IMU, PSD, \newline NTSC cameras \newline (dosimeter, 3D scanner\textsuperscript{*}) 
 & LiDAR, IMU, \newline thermal camera 
 & 3D LiDAR, RGB-D, \newline 360° camera,\newline thermal camera
 & 2× 3D LiDAR, \newline RGB-D cameras, \newline thermal camera, \newline force/torque sensor \\

\hline
\end{tabular}
\label{tab:tracked_robot_comparison}
\end{table*}

Quince \cite{rohmer2010quince} is a compact articulated tracked robot developed for search and rescue operations in confined and hazardous environments. Its design features a main tracked body with four independently driven flippers, providing high mobility on rubble, stairs, and steep inclines. The lightweight carbon-fiber and aluminum chassis (\SI{27}{\kilogram}) ensures structural stiffness and waterproofing, while a low center of gravity enhances stability. The Fukushima redesign \cite{nagatani2011redesign} further strengthened the platform for deployment in the damaged nuclear plant by improving mechanical robustness, power reliability, and lighting, and by adding a 2-DoF manipulator, radiation dosimeter, and a wired/wireless hybrid communication system. These modifications increased the weight to about \SI{50}{\kilogram} but enabled extended operation in complex, radiation-exposed environments.

Karo~\cite{habibian2021design} is a maxi-sized tracked rescue robot featuring a four-flipper chassis, designed for high terrainability on rubble, stairs, and steep slopes. 
A modular aluminum frame with independent, geared DC drives enables climbing inclines of up to \SI{40}{\degree} and crossing gaps of up to \SI{45}{\centi\metre} while maintaining stability. 
A lightweight 7-DoF manipulator (carbon-fiber links, Dynamixel actuators) provides a reach of approximately \SI{1.3}{\metre} with a payload capacity of around \SI{8}{\kilogram}, enabling dexterous tasks such as door and valve operation or object retrieval. 
The robot is equipped with a 2D LiDAR, RGB-D and thermal cameras, an IMU, and gas sensors for mapping, inspection, and victim detection in confined environments.

Our previous platform, Asterix~\cite{asterix2021}, is a compact, autonomy-focused tracked rescue robot designed for operation in unstructured and confined environments. 
The robot features a robust aluminum chassis with four flippers coupled in pairs, providing two active degrees of freedom for stair climbing and step negotiation. 
At a total mass of around \SI{57}{\kilogram}, Asterix reaches approximately \SI{1.1}{\meter/\second} on flat terrain and can overcome steps of up to \SI{30}{\centi\metre} and inclines of up to \SI{45}{\degree}. 
Modular payloads support autonomous navigation and situational awareness, including a rotating 3D LiDAR and multiple RGB-D and 360° cameras. 
A 6-DoF manipulator with a reach of \SI{0.85}{\metre} and a payload capacity of \SI{2.5}{\kilogram} enables dexterous inspection and manipulation tasks such as valve turning and object handling.

In the RoboCup Rescue League~\cite{robocuprescue2015}, robots are tested in arenas with unstructured terrain, obstacles, tight spaces, and objects requiring manipulation, such as doors, valves, and push buttons.
All of the top-performing teams in the years 2022 to 2024 built their own robots, and the design of a tracked robot with a large manipulator and four reconfigurable flippers has proven highly successful in recent years.
However, to the authors' knowledge, none of the top-performing teams have published their design, and only limited information is available from the team description papers\footnote{\href{https://tdp.robocup.org/?tdp_league=tdp-lg-robocuprescue-robot&post_type=tdp_post&s=&tdp_year[]=2022&tdp_year[]=2023&tdp_league[]=tdp-lg-robocuprescue-robot&tdp_team[]=hector-darmstadt&tdp_team[]=irap-robot&tdp_team[]=quix&tdp_team[]=shinobi}{https://tdp.robocup.org/?tdp\_league=tdp-lg-robocuprescue-robot\&post\_type=tdp\_post}}. 
 
The design of \robotName~is greatly inspired by the design of Team Shinobi and Team Quix with four reconfigurable flippers and a large central manipulator, and a successor to our previous platform \prevRobotName~\cite{asterix2021}.
It introduces several novelties, including a high-resolution environment sensing module that captures depth and RGB data, a large manipulator equipped with a force-torque sensor, and a flexible mounting solution for the track profiles, which facilitates further experiments with different profile designs and materials.

\section{LOCOMOTION}
In our previous platform \prevRobotName, the locomotion system was realized using profiles mounted on bike chains.
While this design was very flexible and allowed the testing of different profile designs, the profiles only covered a small part of the robot's footprint, which could lead to the robot getting stuck on unstructured terrain.
In addition, the bike chain design was prone to derailing from the sprocket when turning in unstructured terrain, and the later-added guiding rails to reduce this problem significantly increased the friction for the drive motors, leading to poor efficiency and frequent overheating of the motors.

To address these shortcomings, for \robotName, we have designed a new locomotion system that uses a belt drive covering the majority of the robot's footprint, which is guided by the chassis with low friction losses.
The chassis consists of a very lightweight and strong carbon-fiber reinforced plastics (CFRP) structure glued from \SI{3}{\milli\metre} CFRP plates with \SI{0}{\degree}/\SI{90}{\degree} fiber orientation, and a \SI{3}{\milli\metre} aluminum base plate.
The aluminum base plate aids with thermal regulation for the internal electronics -- as the CFRP plates used in our construction exhibit low thermal conductivity -- and the higher weight lowers the center of mass, which is advantageous during obstacle traversal.
The size of the chassis was chosen such that it fits inside a large suitcase for international transportation.

\begin{figure}
  \centering
  \includegraphics[width=0.48\linewidth]{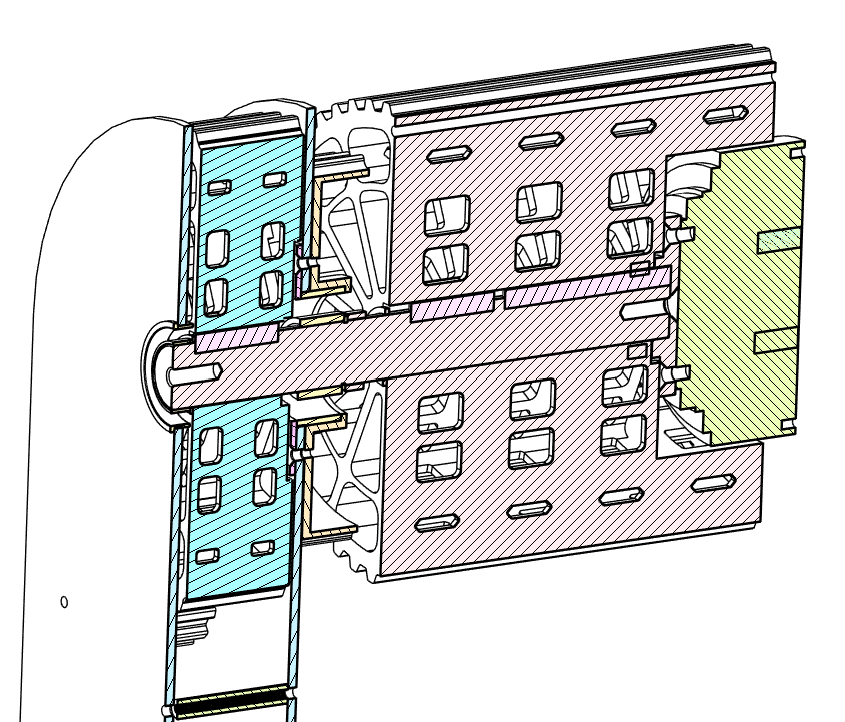}%
  \hspace{0.02\linewidth}%
  \includegraphics[width=0.48\linewidth]{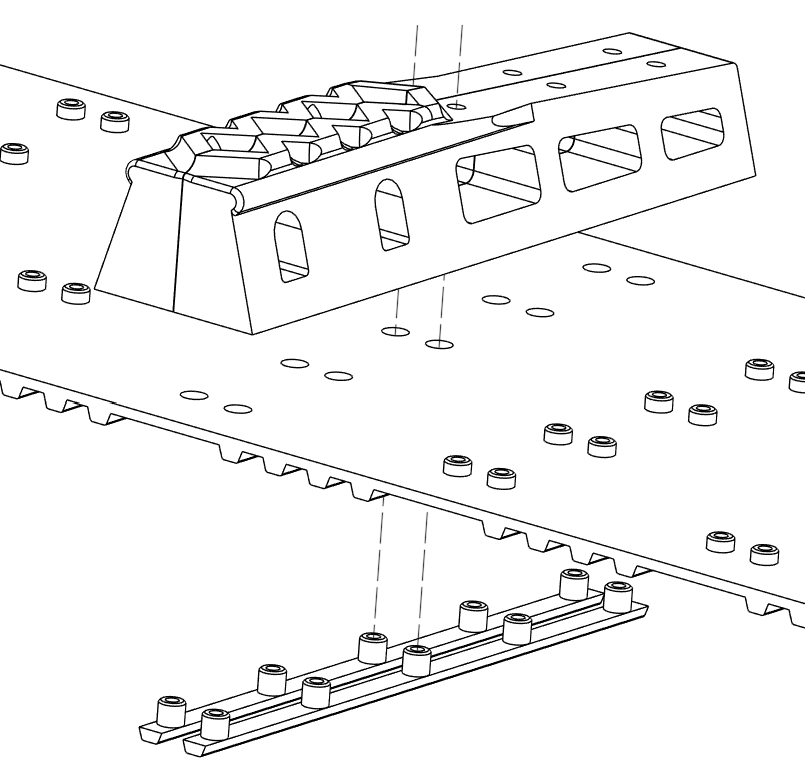}%
  \caption{(Left) Cross-section of the motor, shaft, and pulleys driving the belt. (Right) Profile mounting on the belt using steel tooth inserts with threads.}
  \label{fig:locomotion}
\end{figure}

Four individually reconfigurable flippers driven by Dynamixel H54-200-S500-R servo motors with a 2:1 translation further allow the robot to adapt to unstructured terrain and climb obstacles.
The length was chosen such that it is possible to perform full rotations.
When fully extended, the total length of the robot increases from \SI{72}{\centi\metre} to \SI{128}{\centi\metre}.

As main drive motors, two Unitree A1 motors are mounted to the CFRP chassis on each side.
Each A1 motor can provide up to \Nm{30} for a combined \Nm{60} per side with a maximum speed of 3 rotations per second, resulting in a maximum speed of up to \SI{1.5}{\metre/\second}.
The motor's rotation is transferred to the belt using mounted shafts with slot nut inserts.
Using slot nuts, the shaft rotation is linked to the PETG 3D-printed drive pulleys of the main belt, as well as the flipper pulleys (see Fig. \ref{fig:locomotion}).
While for the main belt with a large contact area, the 3D-printed pulleys exhibit sufficient torque-transmission capacity, the flipper pulleys are reinforced using two \SI{2}{\milli\meter} steel plates per pulley in a 3D-printed sandwich structure.
In addition to the driving pulley and the driven pulley, smaller support pulleys ensure the flipper's belt is pressed flat on obstacles and does not give in.

The belt is a processed industrial PU T10 belt with steel cord.
The length was chosen as the optimal belt length plus \SI{4}{\milli\metre} to allow easy assembly, as we have observed and confirmed with Team Shinobi that the optimal belt length plus \SI{2}{\milli\metre} is difficult to put on the pulleys without a tensioning mechanism.

A key novelty of this belt design is the ability to replace the profiles in the event of wear or to experiment with different profile designs and materials.
Similar approaches exist for tracked robots such as the Telerob Telemax where individual track links can be replaced, as well as large machinery.
Our approach uses a processing method that is commonly used in industrial conveyor belts and thus can be ordered from belt manufacturers at competitive rates.
For this, the belt is modified such that every sixth tooth, two teeth are removed and holes are drilled in the belt\footnote{Note that due to the use of two consecutive inserts, AT10 can not be used as the increased tooth width could lead to jamming.} (see Fig. \ref{fig:locomotion}).
Five holes for the main belt and two for the flipper belts.
In these holes, steel tooth inserts with threads are mounted.
The inserts are used to screw on custom profiles made from PT Flex 70 and Poly 75-70 casting rubber with a shore hardness of A70 for good ground contact and grip, with a PETG inner skeleton for added stability.

The profiles were designed with a V-shape formed by two consecutive profiles to redirect part of the force applied when the profile grips into an obstacle, such as a step, into the belt.
The split into two consecutive profiles allows the profile to move easily over the pulleys.
Additionally, we have added a small round notch at the top to enable the profile to grip into steps.
Arrows on top of the profiles further improve the ground friction on low-friction grounds such as plywood.

\section{Manipulator}
\begin{figure}
  \centering
  \includegraphics[width=0.88\linewidth]{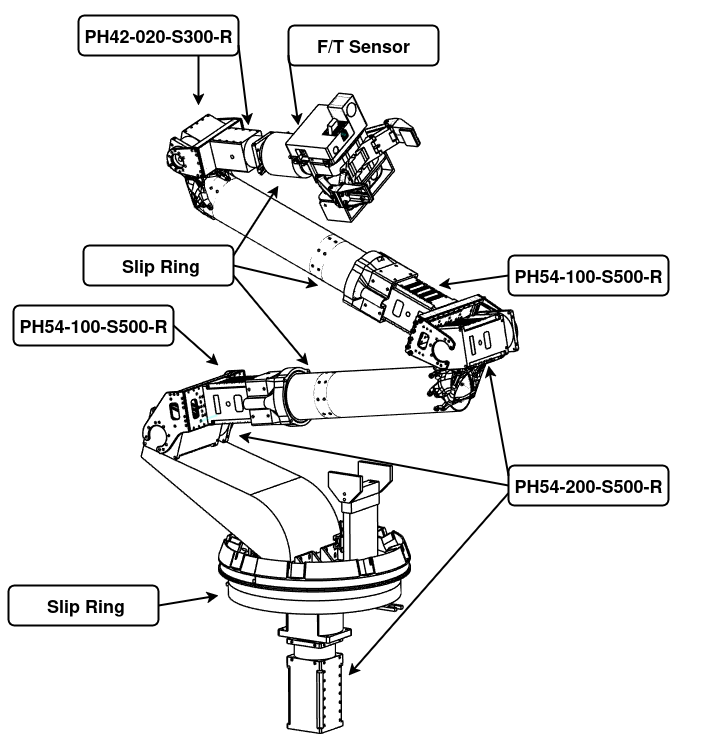}%
  \caption{Overview of the manipulator, highlighting the kinematic chain and the integration of continuous-rotation slip rings within each roll joint. Actuator types are indicated along the chain.}
  \label{fig:manipulator}
\end{figure}

Many disaster scenarios require the inspection or manipulation of objects such as doors, valves, or buttons.
The manipulator is designed to be lightweight, compact when stored, and highly dexterous while integrating visual and force sensing.
As shown in Fig.~\ref{fig:manipulator}, it features seven active joints (eight, including the gripper) with a maximum reach of \SI{1.54}{\metre} and a payload capacity of up to \SI{2.9}{\kilogram} when fully extended.
Seven revolute joints make the arm kinematically redundant for arbitrary end-effector poses, enabling null-space motion and multiple inverse kinematics solutions for the same task frame. 
Centrally mounted on the chassis, it can access objects from any direction and can be magnetically stowed on an armrest during locomotion to reduce sensor occlusion and safeguard the arm.

The base yaw joint is actuated by a Dynamixel PH54-200-S500-R, delivering a maximum torque of \SI{44.7}{\newton\metre}, and is mounted between the base and top plates of the chassis.
The output shaft is supported by a large ball bearing to absorb tilting moments, and a pancake slip ring enables continuous rotation.
The slip ring provides three \SI{24}{\volt} \SI{10}{\ampere} power lines, one \SI{24}{\volt} \SI{5}{\ampere} line, one \SI{5}{\volt} \SI{5}{\ampere} line, an emergency stop line, and a \SI[per-mode=symbol]{100}{\mega\bit} Ethernet connection.
The rotating base also features an LED ring, a microphone array, and a speaker, enabling illumination, communication, and sound source localization.
A fixed curved link extends the reach of the arm by moving the first pitch motor, a Dynamixel PH54-200-S500-R, outward as far as possible without colliding with the flippers in an upright position.

Subsequent links are constructed from cylindrical carbon-fiber-reinforced polymer (CFRP) tubes to maximize stiffness-to-weight ratio.
Each tube houses a through-bore slip ring for internal power and data transmission, while allowing continuous rotation of all roll joints.
The slip rings feature one \SI{24}{\volt} \SI{10}{\ampere} powerline, one \SI{5}{\volt} \SI{5}{\ampere} and \SI{100}{\mega\bit} ethernet.
The third actuator (PH54-100-S500-R, \SI{25.3}{\newton\metre}) connects via such a tube to the second pitch actuator (PH54-200-S500-R), followed by the elbow roll joint using another PH54-100-S500-R actuator.
The second and third pitch actuators are connected to the CFRP tube via 3D-printed titanium brackets, optimized to maximize the joint range of the pitch joints.  
Compact PH42-020-S300-R motors actuate the wrist pitch and roll axes, each providing \SI{5.1}{\newton\metre}. The last roll joint also incorporates a slip ring, ensuring continuous rotation of the gripper.

Mounted on the wrist is an ATI Mini45 force–torque sensor enabling compliant interaction. 
The gripper itself is a Dynamixel RH-P12-RN with passive compliant fingers and a maximum gripping force of \SI{170}{\newton}.
Several sensors are attached to the gripper: an Orbbec Gemini 2 RGBD camera, a Seek Thermal Nano 300, a wide-angle RGB camera, and a high-power LED.
These are connected to a Radxa Zero 3E single-board computer that performs local processing and communicates with the robot’s main computer via Ethernet.
The RGB camera is used for teleoperation, while the RGBD sensor provides 3D data for reconstruction and autonomous manipulation. The thermal camera enables temperature-based inspection in disaster environments.

\section{Electronics}
In this section, we will give a short overview of the sensors for environment perception, the power distribution, the compute for sensor processing, and the connectivity.
\subsection{Sensors}
The robot is equipped with a variety of sensors to provide a high coverage of the environment.
All sensors on the base are mounted symmetrically to maximize coverage and minimize the importance of the driving direction.

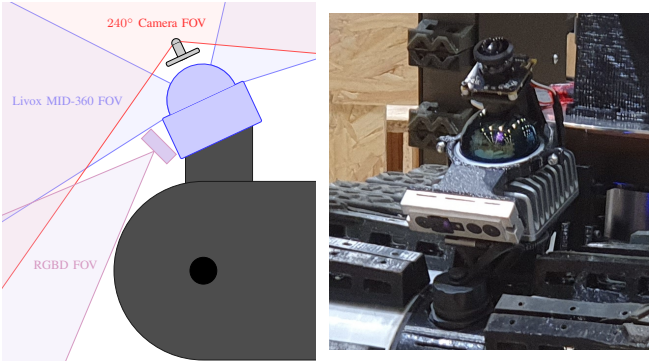
\begin{figure}
  \resizebox{0.49\linewidth}{!}{\begin{tikzpicture}[scale=1, 
    livox/.style={fill=blue!20, draw=blue!80, rounded corners=1pt},
    camera/.style={fill=gray!40, draw=black, rounded corners=1pt},
    pcbborder/.style={draw=gray!80, thick},
    fovlivox/.style={draw=blue!50, fill=blue!20, fill opacity=0.2},
    fovcamera/.style={draw=red!70, fill=red!20, fill opacity=0.2},
    rgbd/.style={draw=red!60!blue!40, fill=red!30!blue!20},
    fovrgbd/.style={draw=red!60!blue!40, fill=red!30!blue!20, fill opacity=0.2},
    chassis/.style={draw=black!90, fill=black!70}
    ]

\def\livoxwidth{2}
\def\livoxheight{1}
\def\gap{0.25}
\def\cameraFOVLength{12}
\def\rgbdFOVLength{12}
\def\livoxFOVLength{12}

\clip (-4, -4.5) rectangle (3, 3.5); %

\draw[chassis] (0.1, -1) rectangle ++(1.5, 2);

\begin{scope}[rotate=25]
    \coordinate (LivoxBase) at (0,0);
    \draw[fovlivox] (LivoxBase) ++(0.2,1) arc[start angle=180, end angle=128, radius=0.8] -- ++(128:\livoxFOVLength) arc[start angle=128, end angle=187, radius=\livoxFOVLength+0.8] -- cycle;
    \draw[fovlivox] (LivoxBase) ++(1.8,1) arc[start angle=0, end angle=52, radius=0.8] -- ++(52:\livoxFOVLength) arc[start angle=52, end angle=-7, radius=\livoxFOVLength+0.8] -- cycle;
    \draw node[blue!50] at (-1.75, 2.25) {\footnotesize Livox MID-360 FOV};

    \draw[livox] (LivoxBase) rectangle ++(2, 1);
    \draw[livox] (LivoxBase) ++(1.8,1) arc[start angle=0, end angle=180, radius=0.8];

    \begin{scope}[shift={(0.6,2)}]
      \draw[fovcamera] (0.4,0.45) -- ++(330:\cameraFOVLength) arc[start angle=-30, end angle=210, radius=\cameraFOVLength] -- cycle;
      \draw node[red!80] at (0.2,1) {\footnotesize $240^\circ$ Camera FOV};

      \draw[camera] (0.3, 0) rectangle ++(0.2, 0.4);
      \draw[camera] (0.5, 0.4) arc[start angle=0, end angle=180, radius=0.1];
      \draw[camera] (0,0) rectangle ++(0.8, 0.1);
    \end{scope}
\end{scope}

\begin{scope}[rotate=45, shift={(-0.3, 0.15)}]
  \draw[fovrgbd] (0,0.4) -- ++(202.15:\rgbdFOVLength) arc[start angle=202.15, end angle=180-22.15, radius=\rgbdFOVLength] -- cycle;
  \draw node[red!60!blue!40] at (-3.2, 0) {\footnotesize RGBD FOV};
  \draw[rgbd] (0, 0) rectangle ++(0.3,0.8);
\end{scope}

\begin{scope}[shift={(-3.5, -4.5)}]
  \draw[chassis] (8, 4) -- (4, 4) arc[start angle=90, end angle=270, radius=2] -- (8, 0) -- cycle;
  \draw[fill=black!100] (4, 2) circle[radius=0.3];
\end{scope}

\end{tikzpicture}}
  \raisebox{0.4em}{\includegraphics[width=0.49\linewidth]{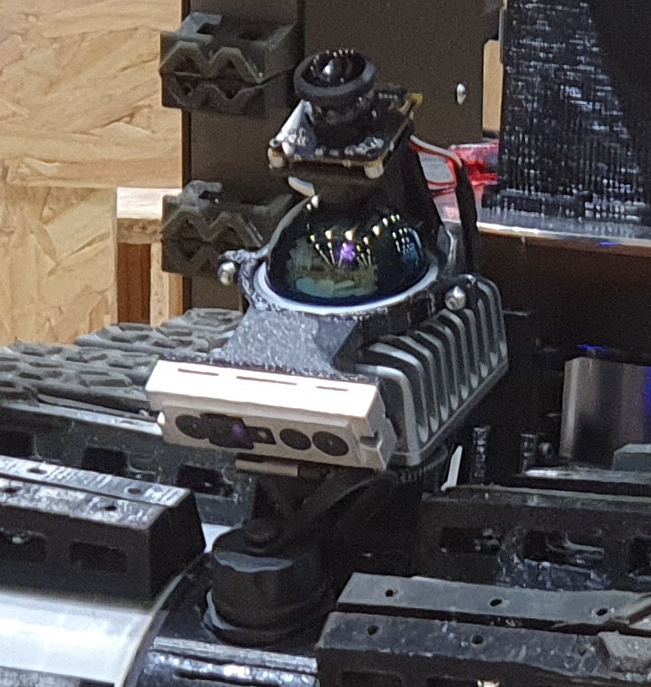}}
  \caption{\robotName's environment perception module with a Livox MID-360 LiDAR, an Orbbec Astra Stereo S U3 RGBD camera, and a 240° wide-angle camera. One module is mounted at each end of the robot.}
  \label{fig:sensors}
\end{figure}

At both ends of the robot, an environment perception module is mounted, consisting of a Livox MID-360 LiDAR with a 240° wide-angle camera on top for colorization and an Orbbec Astra Stereo S U3 RGBD camera to cover the close range in front of the robot (see Fig. \ref{fig:sensors}).

This setup, combined with the Livox's scan pattern, can provide a dense colorized point cloud with a range of up to \SI{70}{\metre} and an accuracy of $\leq$ \SI{2}{\centi\metre} (at \SI{10}{\metre}).

\subsection{Compute \& Connectivity}
The sensor data is processed by a Beelink Mini PC SER9 with an AMD Ryzen 9 AI HX 370 (12C/24T@\SI{5.1}{\giga\hertz}) and 32GB RAM.
This computer is responsible for sensor data processing, base and manipulator control, and navigation.
An Nvidia Jetson Orin AGX with 64GB shared memory handles GPU compute, e.g., neural networks for semantic and object detection, and point cloud processing.

The computers are connected using a switch with 2x \SI{2.5}{\giga\bit} ports for the PCs and 5x \SI{1}{\giga\bit} ports for the sensors and the connection to the SBCs on the manipulator.

Wireless connectivity is provided by a TP-Link EAP225 Outdoor access point powered using Power over Ethernet (PoE) and connected to the switch.
The access point running OpenWRT 24.10 provides access to the robot's internal network and allows for remote control and monitoring of the robot.

\subsection{Power Distribution}
The robot is powered by two battery boxes, each containing two 4S LiPo batteries with a capacity of \SI{6750}{\milli\ampere\hour} connected in series for a total of \SI{210}{\watt\hour} at \SI{30}{\volt} per battery box. It can also be powered via cable from an external \SI{24}{\volt} power supply.

A custom-designed power distribution board (PDB) handles the selection of the input power source and automatically switches between the two \SI{30}{\volt} battery packs or the \SI{24}{\volt} power supply using switches consisting of multiple power MOSFETs, controlled by isolated gate drivers. The entire electronics of the robot is designed to be fully functional at both voltage levels.

The cell voltages and presence of the power supply are monitored by a Teensy 4.0 microcontroller which also controls the switching behavior.

Finally, the PDB provides \SI{5}{\volt} and \SI{12}{\volt} power output at \SI{50}{\watt} to other components, in addition to the power from the batteries or external power supply.

As a safety measure, an additional custom E-Stop board controls the power to the motors using High-Side Power Switches (HSPS).
An STM32 Nucleo L432KC board is used to measure the currents reported by the HSPS and control their enabled pin, completely cutting power to the motors if the E-Stop is active.
The enable pin is routed through hardware buttons on the manipulator by wire, allowing it to shut off the motors without relying on the microcontroller to operate correctly.
To prevent the HSPS overcurrent protection from tripping due to the high switch-on current of the Dynamixel motors, inrush current limiters are installed on the power lines supplying the flippers and the manipulator.

\subsection{Remote E-Stop}
Additionally, a remote E-Stop can be connected and needs to pull the connected pin of the STM32 to HIGH to enable the motors.
We have developed a cost-effective remote E-Stop for experiments with a lower range than commercial solutions (ca. \SI{20}{\metre} obstructed, over \SI{70}{\metre} unobstructed), but at an affordable price of less than 90€.
It is realized using an ESP32 S3 board with a LoRa radio module on both receiver and sender.
The sender is a handheld with display, a twist-to-release button and two additional buttons on the sender.
One button to release the E-Stop and one for a soft E-Stop which does not cut power.
With a base frequency of \SI{20}{\hertz}, the sender transmits the E-Stop state bi-directionally over Bluetooth Low Energy and ESP Now, and additionally at a frequency of approximately \SI{9}{\hertz} unidirectional over LoRa.
If none of the connections received data in the last \SI{300}{\milli\second}, the E-Stop automatically defaults to activate as a fail-safe.
Changes to the E-Stop state are transmitted immediately.

\section{EVALUATION}
The evaluation is separated into the evaluation of the Athena robot and the remote E-Stop.

\begin{figure}
    \centering
    \includegraphics[height=3cm]{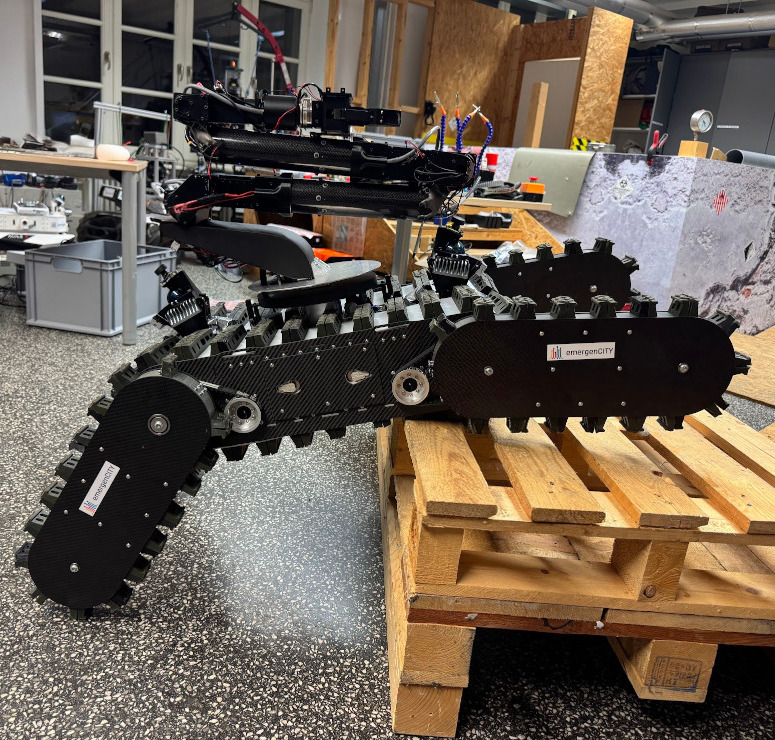}
    \includegraphics[height=3cm]{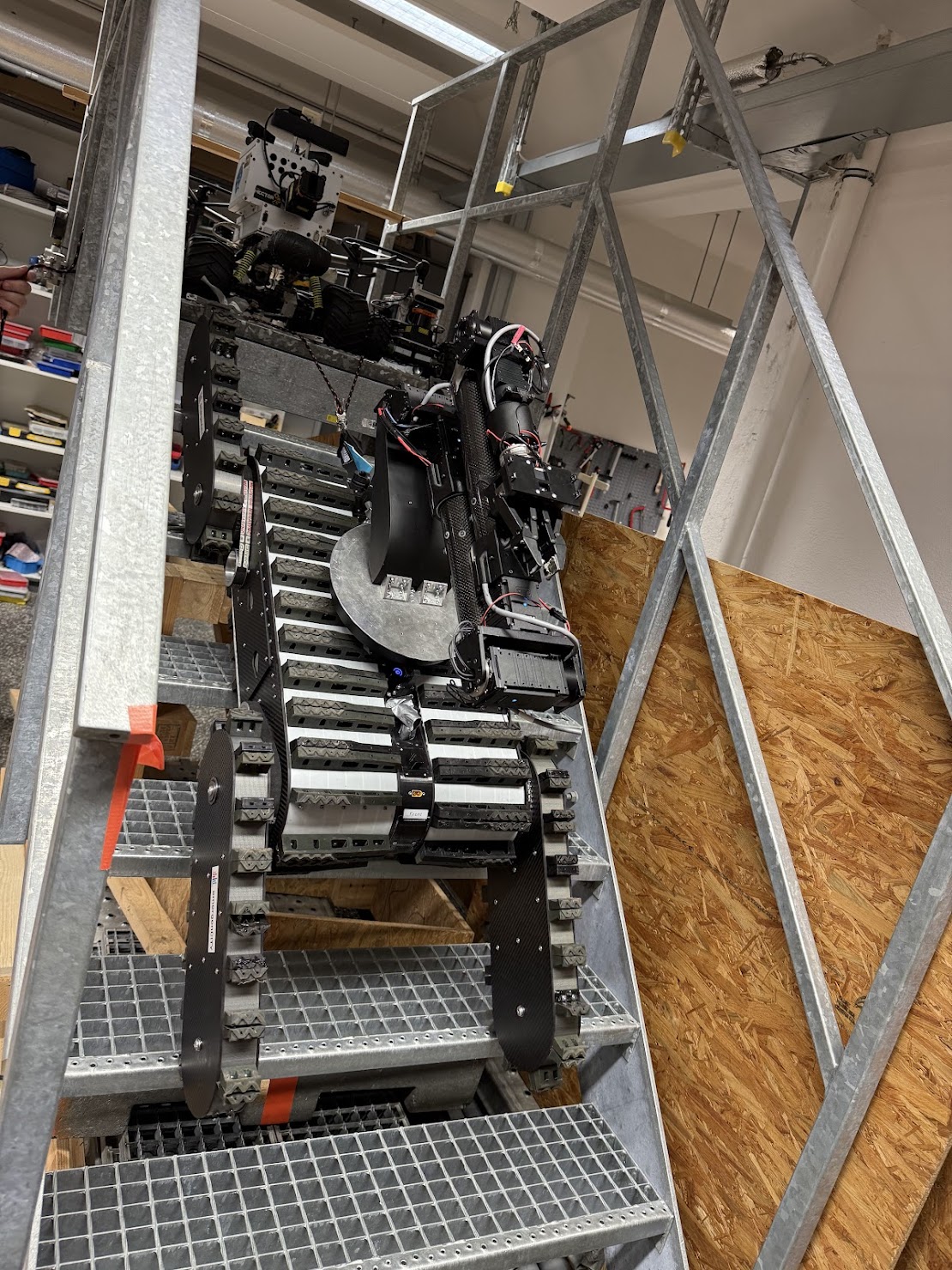}
    \includegraphics[height=3cm]{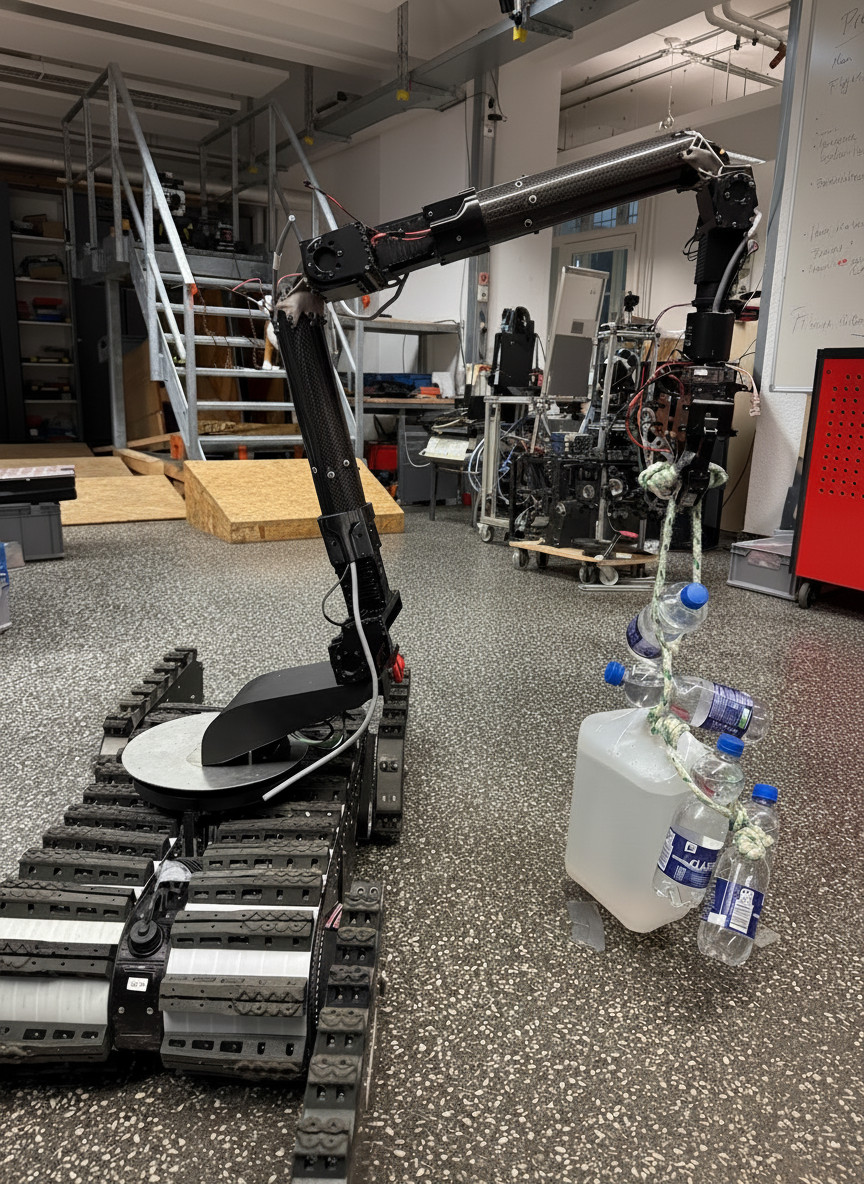}
    \caption{Evaluation scenarios for the maximum step height, application on stairs and the maximum payload of the manipulator.}
\end{figure}

\begin{table*}[ht]
    \centering
    \begin{tabular}{l|c|c|c|c|c}
        Scenario & Line of Sight (\SI{12}{\meter}) & Obstructed (\SI{3}{\meter}) & Stone Wall (\SI{12}{\meter}) & Glass Door (\SI{12}{\meter}) & LoRa Only  \\
        \hline
        Timings & \SI{8+-3}{\milli\second} & \SI{8+-5}{\milli\second} & \SI{33+-29}{\milli\second} & \SI{8+-4}{\milli\second} & \SI{249+-4}{\milli\second} \\
        Maximum & \SI{29}{\milli\second} & \SI{113}{\milli\second} & \SI{128}{\milli\second} & \SI{62}{\milli\second} & \SI{268}{\milli\second}
    \end{tabular}
    \caption{Time in milliseconds from the sender setting the E-Stop state until the power output of the E-Stop board on the robot switches.}
    \label{tab:estop_timings}
\end{table*}

\subsection{Athena}

The terrainability was tested at the RoboCup Rescue League in Brazil, 2025 with continuous stress-tests on the crossing ramps, pallets and stairs with obstacles.
Here, it was identified that while 3D-printed pulleys for the main belt were able to withstand the stress and the motor moments in our test scenarios, the much smaller flipper drive pulleys were unable to resist the motor moments in the stairs scenario when the track was locked.
Here, we reinforced the design using a sandwich structure of a 3D-printed pulley with a laser-cut steel plate to transfer the moments from the shaft to the pulley using multiple screw bolts.
In our first iteration of track profiles, we identified parts made out of Poly 75-70 to be significantly more wear-and-tear-resistant than PT Flex 70 where the majority of profiles made of the latter were partially ripped or completely torn off.
Of the Poly 75-70 profiles only a few profiles were torn off through the screw mounting hole which we will iterate on by adding a steel plate embedded in the profile to reinforce the connection at the screws.

During experiments in our laboratory, we identified the maximum climbable step height using stacked pallets with a total height of \SI{41}{\centi\meter} and the ability to climb stairs of up to \SI{45}{\degree}.
Stairs with steeper inclination were not tested during our experiments.

The maximum payload of the manipulator was evaluated by lifting increasingly heavy weights at the distances \SI{0.5}{\meter}, \SI{1}{\meter} and \SI{1.54}{\meter} until the arm was incapable of lifting and holding the object for at least 20 seconds. At the closest tested distance of \SI{0.5}{\meter} to the base of the manipulator in the robot's center, Athena can lift up to \SI{7.2}{\kilo\gram}.
At a distance of \SI{1}{\meter}, up to \SI{4.8}{\kilo\gram} were lifted.
When the arm is fully extended, it can reach objects as far as \SI{1.54}{\meter} from the center of the robot and can lift up to \SI{2.9}{\kilo\gram}.    

\subsection{Remote E-Stop}
To evaluate the remote E-Stop, we connected the sender to the power output of the E-Stop board using a voltage divider to enable the sender to measure the voltage available at the output of the E-Stop board.
The reliability was evaluated in five scenarios: Obstructed by a pillar at \SI{3}{\meter}, Line of sight at \SI{12}{\meter}, Through closed glass door at \SI{12}{\meter}, Obstructed through building wall at \SI{12}{\meter}, and Line of sight at \SI{12}{\meter} with only LoRa enabled.

For each scenario, the remote E-Stop toggled the E-Stop state one thousand times and measured the time from setting the E-Stop state to disabled until the power was enabled again.
As the data sent has no significant influence on the transmission time, this gives a more precise estimate as it is not influenced by the time it takes for the voltage to dissipate through the voltage divider nor the safety mechanism that would cut off power if no messages are received within \SI{300}{\milli\second}.
As the results in Table \ref{tab:estop_timings} indicate, the E-Stop operates reliably throughout our experiments, and the delay is in the worst case doubling the human reaction time, whereas under normal operating conditions it is within the error margin of the human reaction time.

\section{CONCLUSION}
We presented Athena, a modern open-hardware rescue robotics research platform with strong locomotion and manipulation abilities, and an affordable remote E-Stop solution.
With four individually reconfigurable flippers, exchangeable belt profiles and a strong manipulator, the platform is well-suited for object traversal and manipulation research in challenging terrain.
The high-coverage symmetric sensor solution and high-performance on-board compute enable further research in autonomy and assistance functions as well as environment perception and semantic environment modeling.

\section*{ACKNOWLEDGMENT}
A special thanks to all members of Team Hector who assisted in the building and integration of this platform.

\renewcommand*{\bibfont}{\footnotesize}
\printbibliography

\end{document}